# Fluorescence Image Histology Pattern Transformation using Image Style Transfer

**Mohammadhassan Izadyyazdanabadi[1, 2], Evgenii Belykh[2], Xiaochun Zhao[2], Leandro Borba Moreira[2], Sirin Gandhi[2], Claudio Cavallo[2], Jennifer Eschbacher[3], Peter Nakaji[3], Mark C. Preul[2], and Yezhou Yang[1, *]**

[1]School of Computing, Informatics, and Decision System Engineering, Arizona State University, Tempe, Arizona, USA
[2] The Loyal and Edith Davis Neurosurgery Research Laboratory, Department of Neurosurgery and Department of Neuropathology[3], Barrow Neurological Institute, St. Joseph's Hospital and Medical Center, Phoenix, Arizona, USA

**\* Correspondence:**
Corresponding Author
yz.yang@asu.edu

**Abstract**

     Confocal laser endomicroscopy (CLE) allow on-the-fly in vivo intraoperative imaging in a discreet field of view, especially for brain tumors, rather than extracting tissue for examination ex vivo with conventional light microscopy. Fluorescein sodium-driven CLE imaging is more interactive, rapid, and portable than conventional hematoxylin and eosin (H&E)-staining. However, it has several limitations: CLE images may be contaminated with artifacts (motion, red blood cells, noise), and neuropathologists are mainly trained on colorful stained histology slides like H&E while the CLE images are gray. To improve the diagnostic quality of CLE, we used a micrograph of an H&E slide from a glioma tumor biopsy and image style transfer, a neural network method for integrating the content and style of two images. This was done through minimizing the deviation of the target image from both the content (CLE) and style (H&E) images. The style transferred images were assessed and compared to conventional H&E histology by neurosurgeons and a neuropathologist who then validated the quality enhancement in 100 pairs of original and transformed images. Average reviewers' score on test images showed 84 out of 100 transformed images had fewer artifacts and more noticeable critical structures compared to their original CLE form. By providing images that are more interpretable than the original CLE images and more rapidly acquired than H&E slides, the style transfer method allows a real-time, cellular-level tissue examination using CLE technology that closely resembles the conventional appearance of H&E staining and may yield better diagnostic recognition than original CLE grayscale images.

## 1 Introduction

     Confocal laser endomicroscopy (CLE) is undergoing rigorous assessment for its potential to assist neurosurgeons to examine tissue in situ during brain surgery (1–5). The ability to scan tissue or

surgical resection bed on-the-fly essentially producing optical biopsies, compatibility with different fluorophores, miniature size of the probe and the portability of the system are essential features of this promising technology. Currently, the most frequent technique used for neurosurgical intraoperative diagnosis is examination of frozen section hematoxylin and eosin (H&E)-stained histology slides.

Figure 1 (a) shows an example image from a glioma acquired by CLE (left) and a micrograph of an H&E slide (right), acquired by conventional light microscopy. Although generating CLE images is much faster than H&E slides (1 second per image compared to about 20 minutes per slide), many CLE images may be non-optimal and can be obscured with artifacts including background noise, blur, and red blood cells (6). The histopathological features of gliomas are often more identifiable in the H&E slide images compared to the CLE images generated using nonspecific fluorescent dyes such as fluorescein sodium (FNa). Neuropathologists as well are used to evaluating detailed histoarchitecture colorfully stained with H&E for diagnoses, especially for frozen section biopsies. Fluorescent images from intraoperative neurosurgical application present a new digital gray scale (monochrome) imaging environment to the neuropathologist for diagnosis that may include hundreds of images from one case. Recently, the U.S. FDA has approved a blue laser range CLE system primarily utilizing FNa for use in neurosurgery.

Countervailing these diagnostic and visual deficiencies in CLE images requires a rapid, automated transformation that can: 1) remove the occluding artifacts, and 2) add (amplify) the histological patterns that are difficult to recognize in the CLE images. Finally, this transformation should avoid removing the critical details (e.g., cell structures, shape) or adding unreal patterns to the image, to maintain the integrity of the image content. Such a method may present "transformed" CLE images to the neuropathologist and neurosurgeon that may resemble images based on familiar and standard, even colored, appearances from histology stains, such as H&E.

One method for implementing this transformation could be supervised learning, however, supervised learning requires paired images (from the same object and location) to learn the mapping between the two domains (CLE and H&E). Creating a dataset of colocalized H&E and CLE images is infeasible because of problems in exact co-localization and intrinsic tissue movements, although small, and artifacts introduced during H&E slide preparation. "Image style transfer", first introduced by Gatys et al. (7), is an image transformation technique that blends the content and style of two separate images to produce the target image. This process minimizes the distance between feature maps of the source and target images using a pretrained convolutional neural network (CNN).

In this study, we aimed to remove the inherent occlusions and enhance the structures that were problematical to recognize in CLE images. Essentially, we attempted to make CLE images generated from non-specific FNa application during glioma surgery appear like standard H&E-stained histology and evaluate the accuracy and usefulness. We used the image style transfer method since it extracts abstract features from the CLE and H&E image that are independent of their location in the image and thus can operate on the images that are not from the same location. More details about the image style transfer algorithm and the quality assessment protocol follow in section 2. Our results from a test dataset showed that on average, the diagnostic quality of stylized images was higher than the original CLE images, although there were some cases where the transformed image showed new artifacts.

## 2 Methods

### 2.1 Image Style Transfer

Image style transfer takes a content and style image as input and produces a target image that shows the structures of the content image and the general appearance of the style image. This is achieved through four main components: 1) a pretrained CNN that extracts feature maps from source and target images, 2) quantitative calculation of the content and style representations for source and target images, 3) a loss function to capture the difference between the content and style representation of source and target images, and 4) an optimization algorithm to minimize the loss function. In contrast to CNN supervised learning, where the model parameters are updated to minimize the prediction error, image style transfer modifies the pixel values of the target image to minimize the loss function while the model parameters are fixed.

A 19-layer visual geometry group network (VGG-19), that is pretrained on ImageNet dataset, extracts feature maps from CLE, H&E, and target images. Feature maps in layer "Conv4_2" of VGG-19 are used to calculate the image content representation, and a list of gram matrices from feature maps of five layers ("ReLU1_1", "ReLU2_1", …, "ReLU5_1") are used to calculate the image style representation. To examine the difference between the target and source images, the following loss function was used:

$$Loss_{Total} = \underbrace{\frac{1}{2}\sum(C_{CLE} - C_{Target})^2}_{\text{Content Loss: difference between content representation of the CLE and target image}} + \alpha \times \underbrace{\sum_{i=1}^{5} w^i \times \sum(S^i_{H\&E} - S^i_{Target})^2}_{\text{Style Loss: difference between style representation of the H\&E and target image}}$$

$C_{CLE}$ and $C_{Target}$ are the content representations of the CLE and target image, $S^i_{H\&E}$ and $S^i_{Target}$ are the style representations of the H&E and target image based on the feature maps of the $i^{th}$ layer, and $w^i$ (weight of $i^{th}$ layer in the style representation) equals 0.2. The parameter $\alpha$ determines relative weight of style loss in the total loss and is set to 100. A limited memory optimization algorithm (L-BFGS (8)) minimizes this loss.

For the experiment, 100 CLE images (from a recent study by Martirosyan et al.(5)) were randomly selected from 15 subjects with glioma tumors as content images. A single micrograph of an H&E slide from a glioma tumor biopsy of a different patient (not one of the 15 subjects) was used as the style image (Figure 1 (a), right). For each CLE image, the optimization process was run for 1600 iterations and the target image was saved for evaluation and referred to as the "stylized image" in the following sections.

## 2.2 Evaluation

Although the stylized images presented the same histological patterns as H&E images and seemed to contain similar structures to those present in the corresponding original CLE images, a quantitative image quality assessment was performed to rigorously evaluate the stylized images. Five neurosurgeons independently assessed the diagnostic quality of the 100 pairs of original and stylized CLE images. For each pair, the reviewers sought to examine two properties in each stylized image and provided a score for each property based on their evaluation: 1) whether the stylization process removed any critical structures (negative impact) or artifacts (positive impact) that were present in the original CLE image, and 2) whether the stylization process added new structures that were not present (negative impact) or were difficult to notice (positive impact) in the original CLE image. The scores are between 0 and 6 with the following annotations: 0, extreme negative impact; 1, moderate negative impact; 2, slight negative impact; 3, no significant impact; 4, slight positive impact; 5, moderate positive impact; and 6, extreme positive impact. (Further information and instructions about the quality assessment survey is available in the supplementary materials section.)

Since the physicians were more familiar with the H&E than the CLE images, it was possible that the reviewers would overestimate the quality of stylized images merely due to their color resemblance to H&E images (during style transfer the color of CLE image is also changed to pink and purple). To explore how the reviewers' scores would change if the stylized images were presented in a different color other than the pink and purple (the common color for H&E images), the stylized images were processed in four different ways: I) 25 stylized images were converted into gray-scale images (averaging the three red, green, and blue channels), II) 25 stylized images were color-coded in green (first converted the image to gray-scale and then set red and blue channels to zero), III) 25 stylized images were color-coded as red (similar approach), and IV) 25 stylized images were used without any further changes (intact H&E). Since there are too many structures in each CLE image, and to examine the images more precisely, we used the center-crop of each original CLE and its stylized version for evaluation. Figure 1 (b) shows some example stylized images used for evaluation.

## 3    Results and Analysis

Figure 2 (a) shows a histogram of all reviewers' scores for the removed artifacts (blue bars) and added structures (orange bars) in the stylized images with different colors. Overall, the number of stylized CLE images that have higher diagnostic quality than the original images (score greater than 3) was significantly larger than those with equal or lower diagnostic quality for both removed artifacts and added structures scores (one-way chi square test p-value<0.001). Results from stylized images that were color-coded (gray, green, red) showed the same trend for the added structures scores, indicating that the improvement was not just because of color resemblance.

There was significant difference between how much the model added structures and removed artifacts. For all the color-coded and intact stylized images, the average of added structures scores was larger than the removed artifacts scores (t-test p-value <0.001). This suggests that the model was better at enhancing the structures that were challenging to recognize than removing the artifacts.

Figure 2 (b) shows the frequency of different combinations of scores for removed artifacts and added structures in an intensity map. Each block represents how many times a rater scored an image with the corresponding values on the x (improvement by added structures) and y (improvement by removed artifacts) axes for that block. The most frequent incident across all the stylized images is the coordinates (5,4), which means moderately adding structures and slightly removing artifacts, followed by (5,5) meaning moderately adding structures and removing artifacts. Although the intensity maps derived from different color-coded images were not exactly similar, the most frequent combination in each group still indicated positive impact in both properties. The most frequent combination of scores, for each of the color-coded images, was as follows: gray = (5,4), green = (5,5), red = (5,4), and intact = (5,4).

As a further analysis, we counted the number of images that had an average score of below 3 to see how often the algorithm removed critical structures or added artifacts that were misleading to the neurosurgeons. From the 100 tested images, 3 images had only critical structures removed, 4 images had only artifacts or unreal added structures, and 2 images had both artifacts added and critical structures removed. On the contrary, 84 images showed improved diagnostic quality through both removed artifacts and added structures that were hard to recognize, 6 images had only artifacts removed, and 5 images had only critical structures added. Figure 1 (b) shows some example stylized images with improved quality compared to the original CLE, and Figure 2 (c) shows two stylized images with decreased diagnostic quality through removed critical structures (top) and added artifacts (bottom).

4       Conclusions

In this study, image style transfer was applied to CLE images from gliomas to enhance their diagnostic quality. Style transfer with an H&E-stained slide image had an overall positive impact on the diagnostic quality of CLE images. The improvement was not solely because of the colorization of CLE images; even the stylized images that were converted to gray, red, and green, reported improved diagnostic quality compared to the original CLE images. Employment of more specific clinical tasks to explore the advantage of stylization in diagnosing gliomas and other tumor types is underway based on this preliminary success. The fact that the style transfer is based on permanent histology H&E, provides an intraoperative advantage. Initial pathology diagnosis for brain tumor surgery is usually based on frozen section histology, with formal diagnosis awaiting the inspection of permanent histology slides requiring one to several days. The style transfer is based on rapidly acquired, on-the-fly (i.e., real time) in vivo intraoperative CLE images that most resemble the permanent histology; therefore, it is a significant advantage for interpretation. Frozen section histology often involves freezing artifact, cutting problems, and may have inconsistent staining for histological characteristics that are important for diagnosis. Style transferred CLE images are then most comparable to the permanent histology, and may be even better because CLE is imaging live tissue without such architectural disturbance.

In the future, application of more advanced methods for transferring patterns in the histology slides to the CLE images will be used to improve their interpretability. Because of the high number of CLE images acquired during a single case, style transfer could add value to such fluorescence images and allows for computer-aided techniques to play a meaningful, convenient, and efficient role to aid the neurosurgeon and neuropathologist in analysis of CLE images and to more rapidly determine diagnosis.

**Acknowledgments**

This research was supported by the Newsome Chair in Neurosurgery Research held by Dr. Preul and by the Barrow Neurological Foundation. Dr. Belykh received scholarship support SP-2240.2018.4.

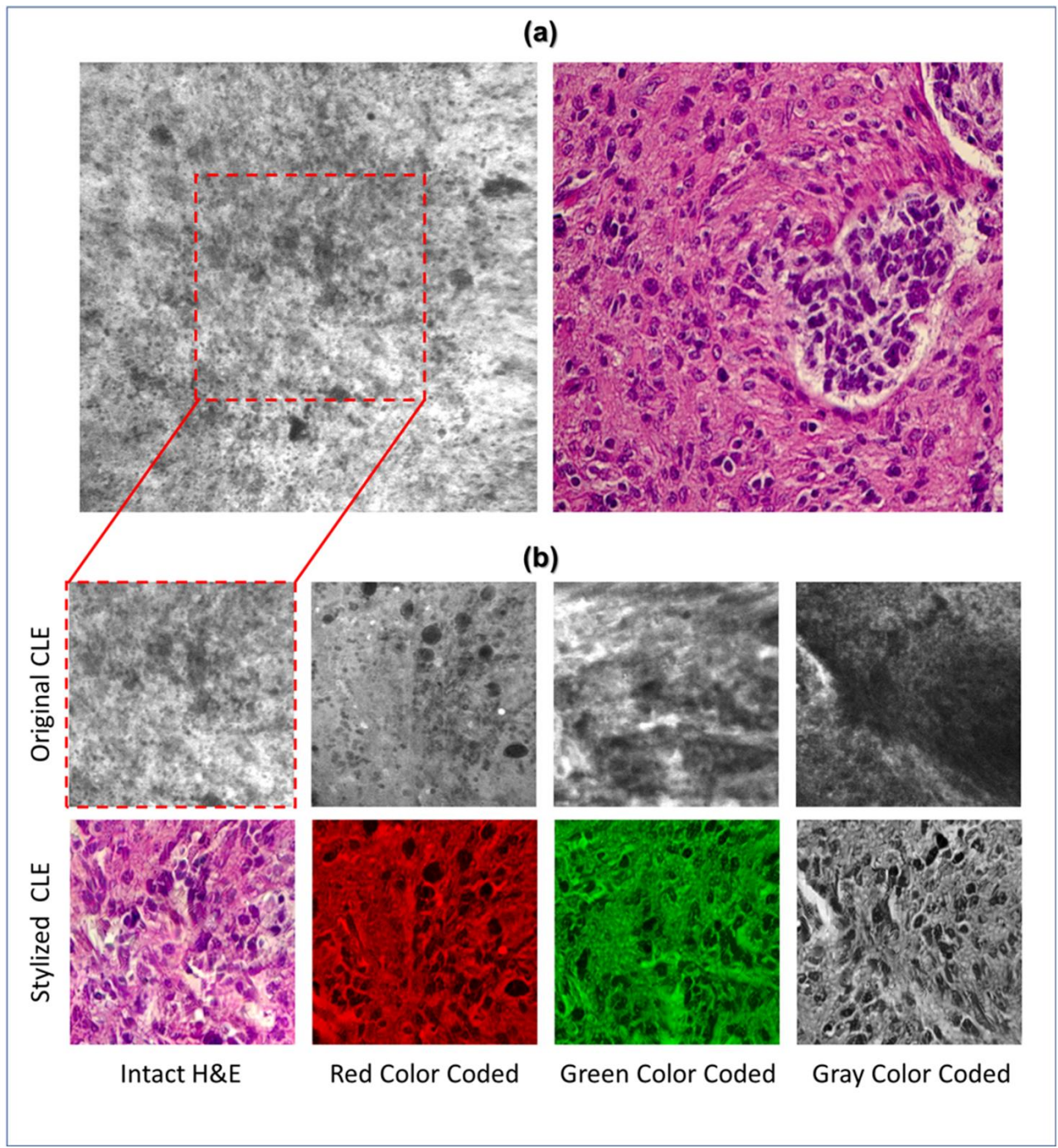

**Figure 1. (a)** Representative CLE (Optiscan 5.1, Optiscan Pty., Ltd.) and H&E images from glioma tumors. **(b)** Original and stylized CLE images from glioma tumors, in 4 color coding: gray, green, red, intact H&E.

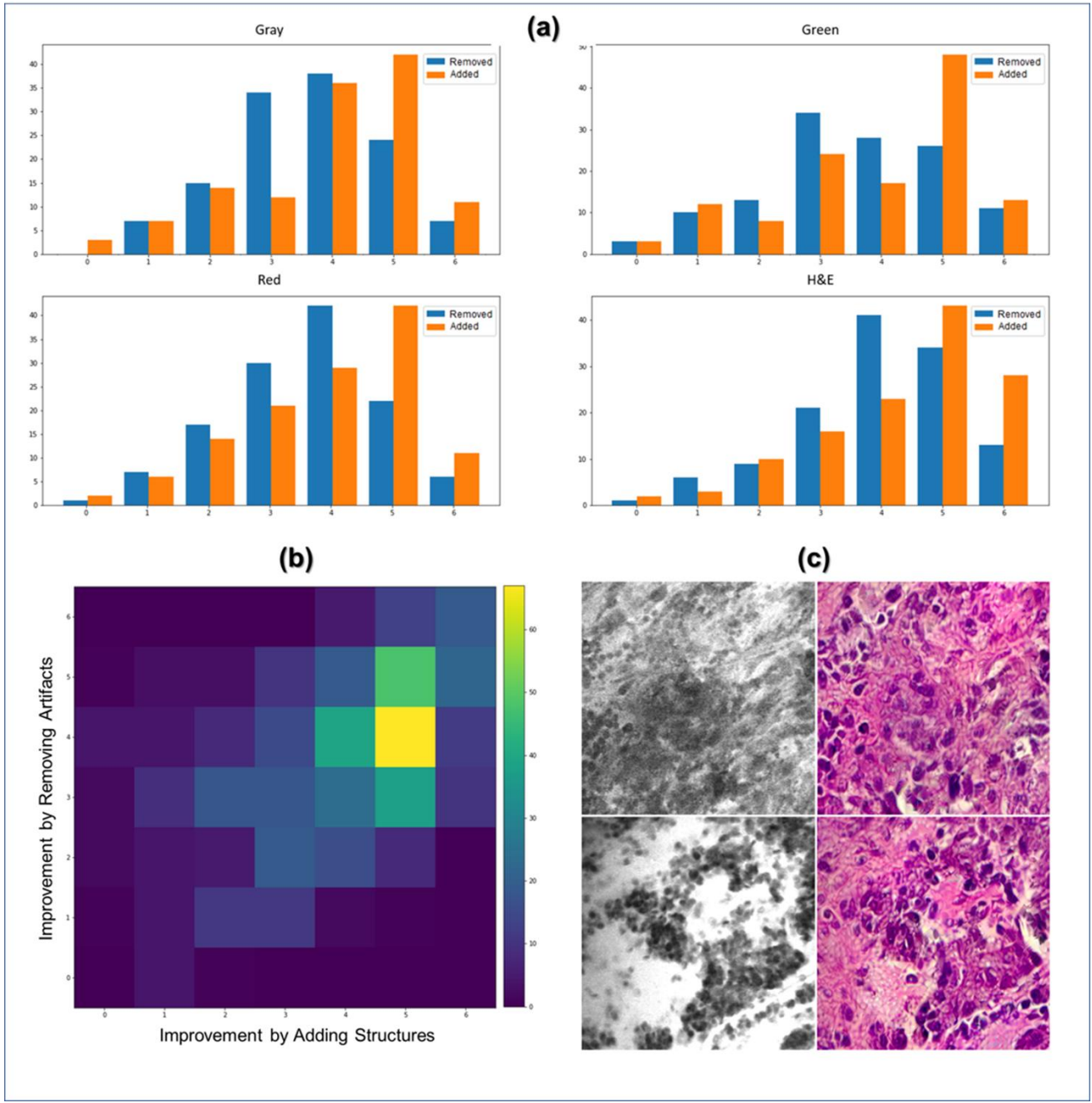

**Figure 2.** **(a)** Histogram of scores for added structures and removed artifacts from different color-coded images. **(b)** An intensity map showing the frequency of different combinations of scores for adding structures (x axis) and removing artifacts (y axis). **(c)** Two example images that the stylization process removed some critical details (top) or added unreal structures (bottom).

# *Supplementary Material*

## Scoring protocol for quality assessment

| Score | Description |
|---|---|
| 0 | Negative* impact; Severe structures are removed |
| 1 | Negative impact; Moderate structures are removed |
| 2 | Negative impact; Slight structures are removed |
| 3 | No significant structures are removed |
| 4 | Positive** impact; Slight artifacts are removed |
| 5 | Positive impact; Moderate artifacts are removed |
| 6 | Positive impact; Severe artifacts are removed |

\* **Negative impact** denotes that the transformed image has a lower diagnostic quality than the original image. (e.g. removing cells or other preexisting diagnostic features)

\*\* **Positive impact** denotes that the transformed image has a higher diagnostic quality than the original image. (e.g. less artifacts)

| Score | Description |
|---|---|
| 0 | Negative* impact; Severe artifacts are added |
| 1 | Negative impact; Moderate artifacts are added |
| 2 | Negative impact; Slight artifacts are added |
| 3 | No significant structures are added |
| 4 | Positive** impact; Slight structures are added |
| 5 | Positive impact; Moderate structures are added |
| 6 | Positive impact; Severe structures are added |

\* **Negative impact** denotes that the transformed image has a lower diagnostic quality than the original image. (e.g. hallucinating cells, misleading structures, and artifacts)

\*\* **Positive impact** denotes that the transformed image has a higher diagnostic quality than the original image. (e.g. highlighting cells or other structures that were hard to notice)